\documentclass[runningheads]{llncs}
\usepackage{graphicx}
\usepackage{amsmath,amssymb} 
\usepackage{cite}
\usepackage{color}
\usepackage{subcaption}
\usepackage{hyperref}
\usepackage{longtable}
\usepackage[width=122mm,left=12mm,paperwidth=146mm,height=193mm,top=12mm,paperheight=217mm]{geometry}

\begin{document}
\newcommand*\samethanks[1][\value{footnote}]{\footnotemark[#1]}
\pagestyle{headings}
\mainmatter
\def\ECCV16SubNumber{3}  

\title{Bi-modal First Impressions Recognition using Temporally Ordered Deep Audio and Stochastic Visual Features} 

\titlerunning{Bi-modal First Impressions - Deep Audio and Stochastic Visual Features} 

\authorrunning{Arulkumar Subramaniam, Vismay Patel et. al}

\author{Arulkumar Subramaniam\thanks{Authors contributed equally}, Vismay Patel\samethanks, Ashish Mishra, \\Prashanth Balasubramanian, Anurag Mittal}
\institute{Department of Computer Science and Engineering,\\ Indian Institute of Technology Madras\\\texttt{\{aruls, vismay, mishra, bprash, amittal\}@cse.iitm.ac.in}}

\maketitle

\begin{abstract}
We propose a novel approach for First Impressions Recognition in terms of the Big Five personality-traits from short videos. The Big Five personality traits is a model to describe human personality using five broad categories: Extraversion, Agreeableness, Conscientiousness, Neuroticism and Openness. We train two bi-modal end-to-end deep neural network architectures using temporally ordered audio and novel stochastic visual features from few frames, without over-fitting. We empirically show that the trained models perform exceptionally well, even after training from a small sub-portions of inputs. Our method is evaluated in ChaLearn LAP 2016 Apparent Personality Analysis (APA) competition using ChaLearn LAP APA2016 dataset and achieved excellent performance.
\end{abstract}

\section{Introduction}

A ``First Impression" is the event when a person encounters another person and forms a mental image about the person \cite{wikifirstimpress}. Here the mental image can be based on lot of characteristics such as facial expressions, action, physical appearance, the way of interaction, body language, etc. According to research in Psychology \cite{willis2006first}, the first impressions are formed even with a limited exposure (as less as 100ms) to unfamiliar faces. Forming a first impression is usually done in terms of Personality-traits recognition. Determining Personality-traits automatically will be helpful in human resourcing, recruitment process. An automatic analysis of Personality-traits will help people to train themselves. 

The problem can be represented as in Table \ref{inputoutput}. A short video with a person's interview is given as input and the output is expected to be 5 fractional values in the range [0, 1] representing Extraversion, Agreeableness, Conscientiousness, Neuroticism, Openness. These five are collectively known as the ``Big-Five personality-traits".


There has not been much work in literature for First-impressions recognition, though the researchers have explored Emotion recognition\cite{cowie2001emotion,cohen2000emotion,cohen2003facial,kim2013deep,Kessous,Kim}, a related area in terms of the type of problem and features (hand-crafted features as well as deep features) used.
There are many ways, people express their emotions, among which facial expressions are the most useful\cite{cowie2001emotion,cohen2000emotion,cohen2003facial,kim2013deep}. Cohen et. al \cite{cohen2000emotion} used HMM based models to categorize the emotions in a video into six types: (1)happy, (2)angry, (3)surprise, (4)disgust, (5)fear, (6)sad. Their extended work\cite{cohen2003facial} in multilevel HMM performed automatic segmentation and recognition from a continuous signal. Xiaowei Zhao et. al \cite{Kim} proposed iterative Multi-Output Random Forests for face analysis in images using a combination of three tasks namely Facial landmark detection, Head pose estimation and Facial expression recognition. Deep features have also been used for facial analysis. Javier G. Razuri et. al \cite{David} have extracted features from regions around eyes and mouth for recognizing the human emotions. Their idea was that information related to emotions could be captured by tracking the expressions around eyes and mouth region. The extracted features are then input into a feed-forward neural network trained by back-propagation for classification of emotions.

Although, facial expressions form an important cue, they alone are not sufficient to recognize emotions effectively. Loic et. al \cite{Kessous} used facial expressions, gestures and acoustic analysis of speech based features. In their work, they have used a Bayesian classifier to recognize one of the eight types of emotions  (Anger, Despair, Interest, Pleasure, Sadness, Irritation, Joy and Pride). They presented uni-modal (trained separately with all three types of features), bi-modal (combine two modes together) and multi-modal (combine all three modes together). Among all combinations, they observed that multi-modal based classification yielded the best performance.

We propose two end-to-end trained deep learning models that use audio features and face images for recognizing first impressions. In the first model, we propose a Volumetric (3D) convolution based deep neural network for determining personality-traits. 3D convolution was also used by Ji et. al \cite{ji3dconv}, although for the task of action recognition from videos of unconstrained settings. In the second model, we formulate an LSTM(Long Short Term Memory) based deep neural network for learning temporal patterns in the audio and visual features. Both the models concatenate the features extracted from audio and visual data in a later stage.  This is in spirit of the observations made in some studies \cite{Kessous} that multi-modal classification yields superior performance. 

Our contribution in this paper is two-fold.  First, mining temporal patterns in audio and visual features is an important cue for recognizing first impressions effectively.  Secondly, such patterns can be mined from a few frames selected in a stochastic manner rather than the complete video, and still predict the first impressions with good accuracy. The proposed methods have been ranked second on the ChaLearn LAP APA2016 challenge(first round)\cite{chalearn1stround1stimpressions}. 

This paper is organized as follows.  In Section \ref{sec:methodology}, we describe the two models in detail and the steps followed to prepare the input data and features for the models.  Section \ref{sec:stochastic_training} describes the novel stochastic method of training and testing the networks.  In Section \ref{sec:experiments_results}, we discuss the Apparent Personality Analysis 2016: First Impressions Dataset, the evaluation protocol, the implementation details and the experimental results obtained in two phases of the competition.  Section \ref{sec:conclusions} concludes the paper providing future direction for the work.

\begin{longtable}{ c| c }
	\multicolumn{1}{c}{Input} &
	\multicolumn{1}{c}{Target} \\\hline
 \includegraphics[scale=0.15]{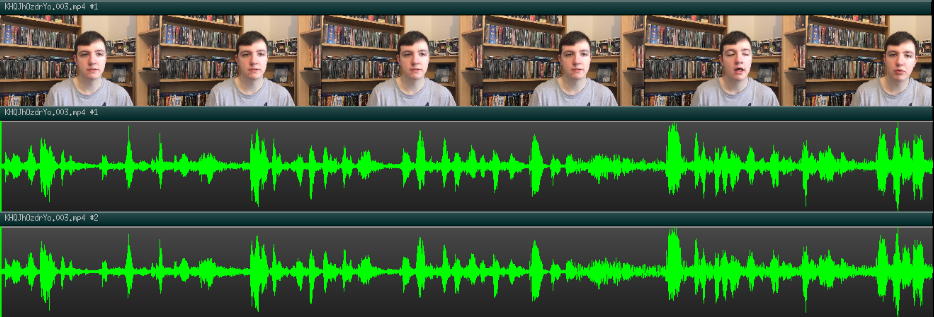} &  \includegraphics[scale=0.15]{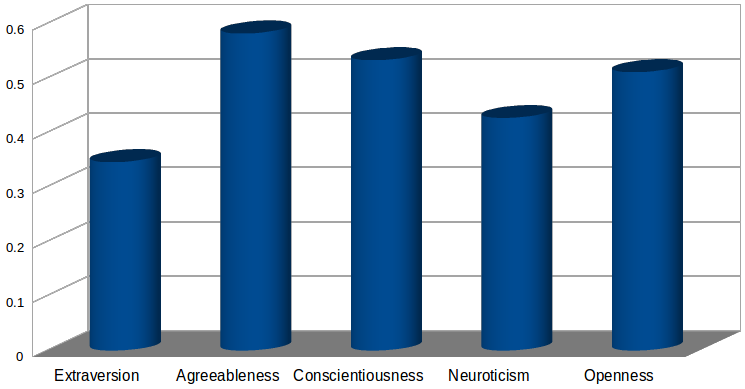} \\ \hline\hline
 \includegraphics[scale=0.15]{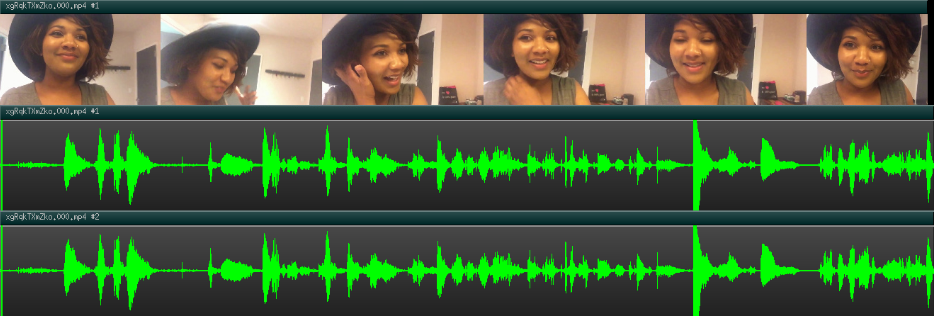} &  \includegraphics[scale=0.15]{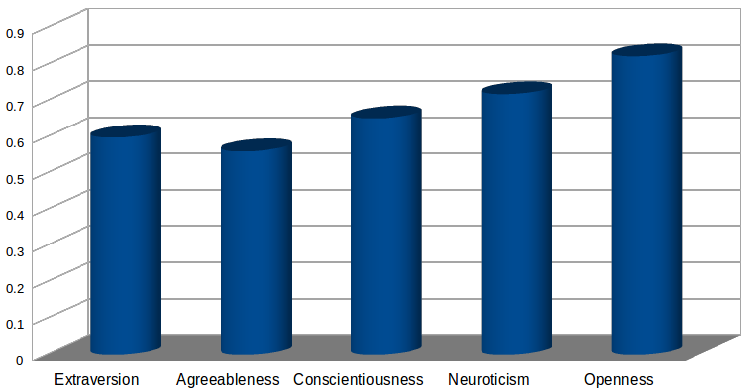} \\ \hline\hline
 \caption{Example of Input and Target. Input is the raw video containing a person's interview \& output will be the predicted personality-traits values. \label{inputoutput}}
\end{longtable}

\section{Methodology}
\label{sec:methodology}
We propose two bi-modal deep neural network architectures that have two branches, one for encoding audio features and the other for visual features. Inputs to both the audio and visual branches of the model are generated after pre-processing the raw video data. Features extracted from both the branches are fused in a later stage of the model, while the complete network is trained end-to-end. In this section, we describe the pre-processing that was performed on the data and the architecture of models in detail.

\begin{figure}
\centering
\includegraphics[width=0.9\textwidth,height=0.3\textheight]{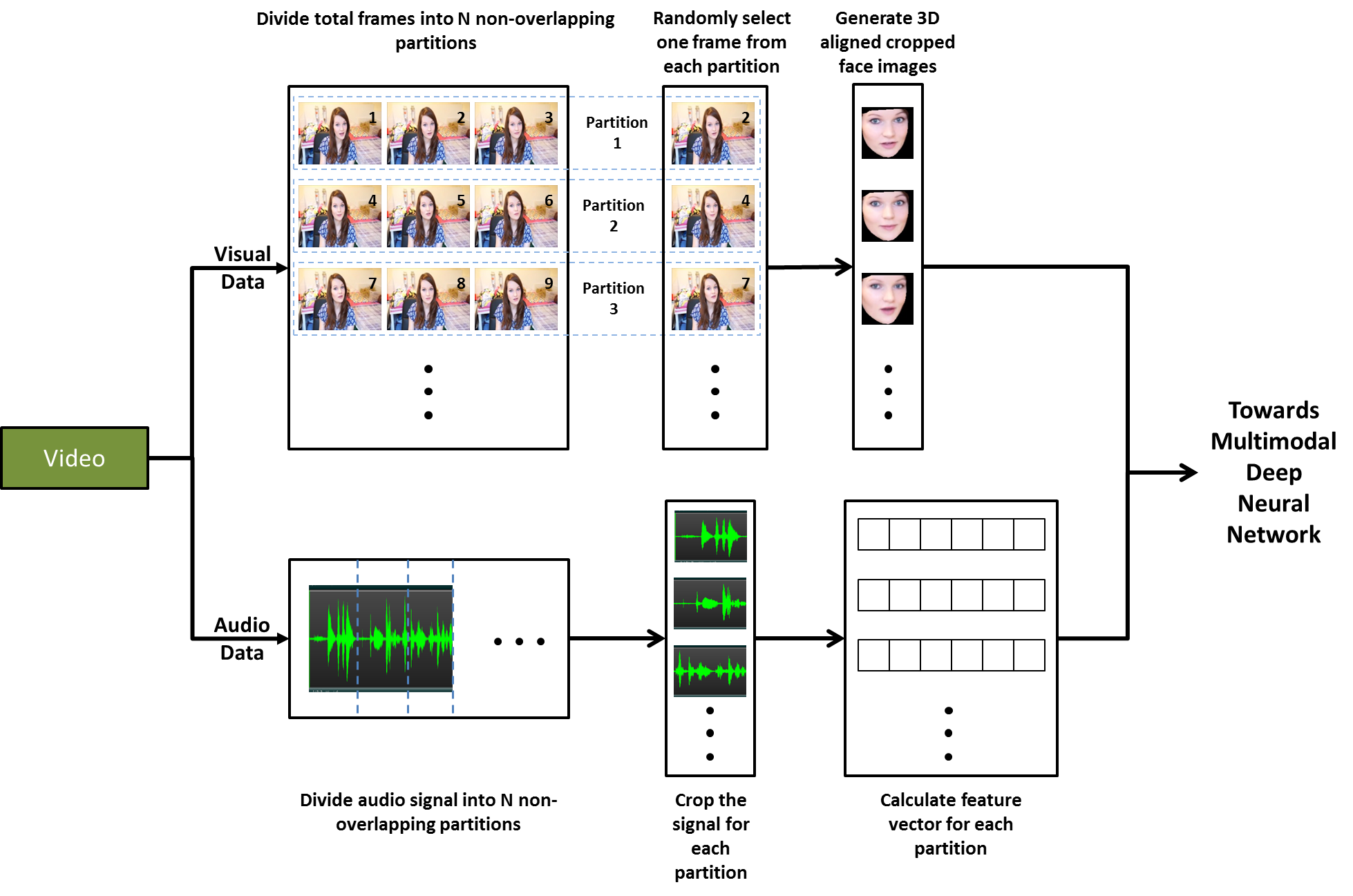}
\caption{Data pre-processing pipeline, where the face aligned images are extracted from image frames and spectral audio features are extracted from audio data.\label{preprocess}}
\end{figure}

\subsection{Audio data pre-processing}
Given a video, we extract its audio component and split the audio component into N non-overlapping partitions as shown in figure \ref{preprocess}. From each individual partition, we extract ``mean and standard deviation" of certain properties (table \ref{audiofeats}) of audio signal. We use an open-source python based audio processing library called pyAudioAnalysis \cite{giannakopoulos2015pyaudioanalysis,pyaudioanalysis} for this purpose. The hand-crafted features are of 68 dimensions, which includes the mean and standard deviation of the following attributes:


  \begin{longtable}{ l | p{9cm} }
	\multicolumn{1}{l}{\textbf{Attribute Name}} &
	\multicolumn{1}{l}{\textbf{Description}} \\
    \hline
    Zero Crossing Rate &  The rate of sign-changes of the signal during the duration  of a particular frame\\\hline
   Energy &	The sum of squares of the signal values, normalized by the respective frame length. \\\hline
	Entropy of Energy &	The entropy of sub-frames' normalized energies. It can be interpreted as a measure of abrupt changes.\\\hline
	Spectral Centroid &	The centre of gravity of the spectrum.\\\hline
	Spectral Spread	& The second central moment of the spectrum.\\\hline
	Spectral Entropy &	Entropy of the normalized spectral energies for a set of sub-frames.\\\hline
    Spectral Flux &	The squared difference between the normalized magnitudes of the spectra of the two successive frames.\\\hline
    Spectral Rolloff &	The frequency below which 90\% of the magnitude distribution of the spectrum is concentrated.\\\hline
	 MFCCs &	Mel Frequency Cepstral Coefficients form a cepstral representation where the frequency bands are not linear but distributed according to the mel-scale.\\\hline
	Chroma Vector	& A 12-element representation of the spectral energy where the bins represent the 12 equal-tempered pitch classes of western-type music (semitone spacing).\\\hline
    Chroma Deviation	& The standard deviation of the 12 chroma coefficients.\\\hline
    
    \hline
    \caption{Audio features extracted using pyAudioAnalysis \cite{pyaudiofeatures} \label{audiofeats}}
  \end{longtable}

\subsection{Visual data pre-processing}
The visual processing branch of the model takes as input, a set of 'N' 3D aligned segmented face images. We segment the face images to prevent the background from affecting the predictions, which should rather depend only on the features of the face (gaze direction, movements of eye, lips, etc).  We use facial landmark detection and tracking to segment the faces. The landmark points are then aligned to fixed locations, which give us segmented face images that have also been aligned.
We use an open-sourced C++ library OpenFace\cite{baltru2016openface,openface} for all the visual pre-processing tasks.

\subsection{Model Architecture}
\begin{figure}[!th]
    \begin{subfigure}[b]{0.5\textwidth}
	\includegraphics[width = 0.9\textwidth, height=0.6\textheight]{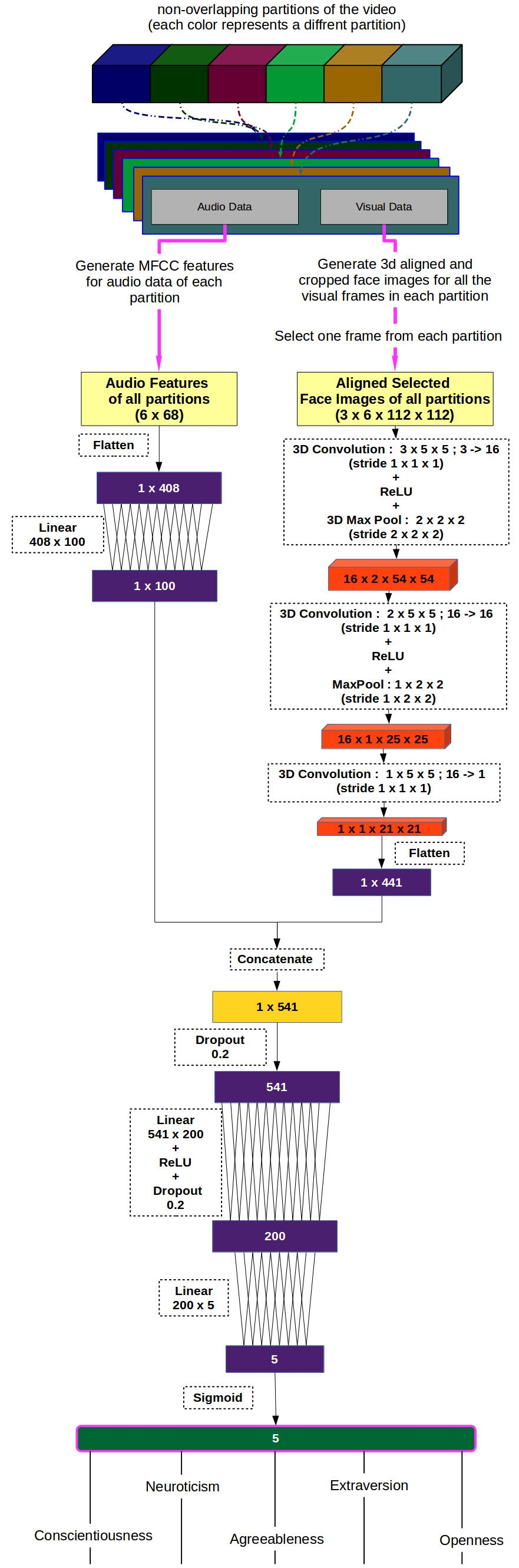}
	\caption{Bi-modal Volumetric Convolutional Neural Network architecture\label{multimodalconv}}
    \end{subfigure}
	\begin{subfigure}[b]{0.5\textwidth}
	\includegraphics[width = 0.9\textwidth, height=0.6\textheight]{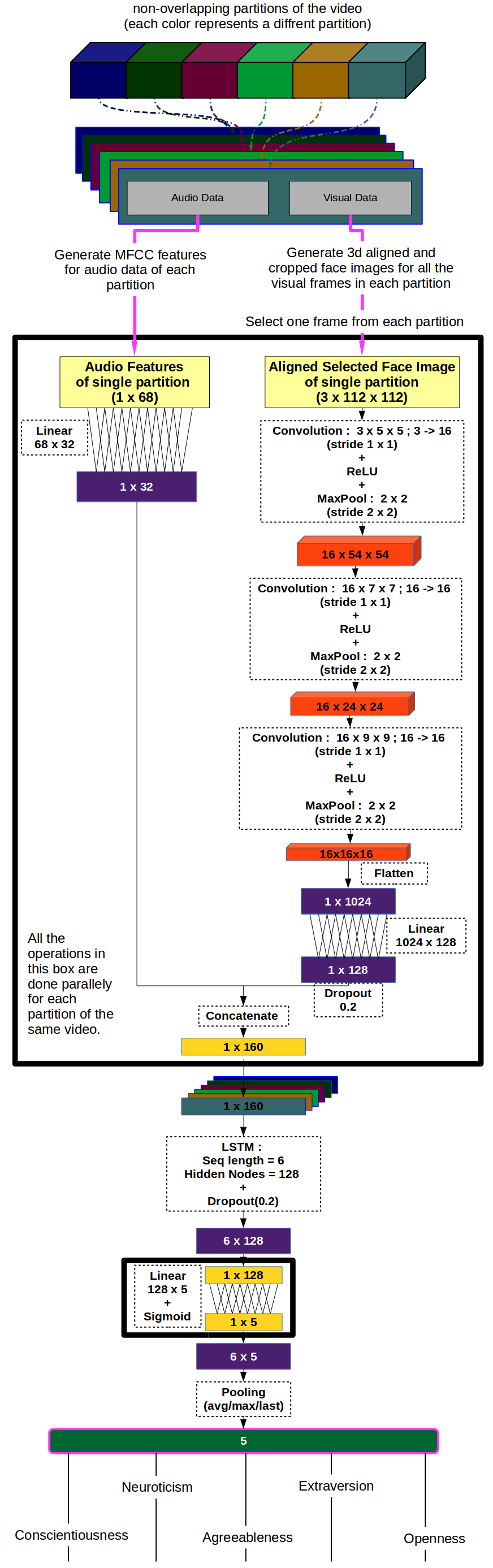}
	\caption{Bi-modal LSTM Neural Network architecture\label{multimodallstm}}
    \end{subfigure}
    \caption{Model Architecture Diagram\label{modelarch}}
\end{figure}

We propose two models in our work. The models are shown in figure \ref{multimodalconv} and \ref{multimodallstm} respectively. We divide each video into N non-overlapping partitions. From each of the N partitions, both audio and visual features are extracted (figure \ref{preprocess}) and used as inputs to the models. Here, only the inter-partition variations are learned as temporal patterns, while the intra-partition variations are ignored. We do so, to handle redundancy in consecutive frames especially in high fps videos. As we can see in figures \ref{convpipe} and \ref{lstmpipe}, the audio and visual features from each block are passed through consecutive layers of neural network. Now, in our first model, the temporal patterns across the N sequential partitions are learned using a 3D convolution module. While in the second model, we use an LSTM to learn the temporal patterns across the partitions. The kernel sizes and stride information are available in the figure \ref{modelarch}. By empirical analysis, we fixed N as 6.
\subsubsection{Volumetric (3D) convolution model:} 
\begin{figure}
\centering
\includegraphics[height=0.25\textheight]{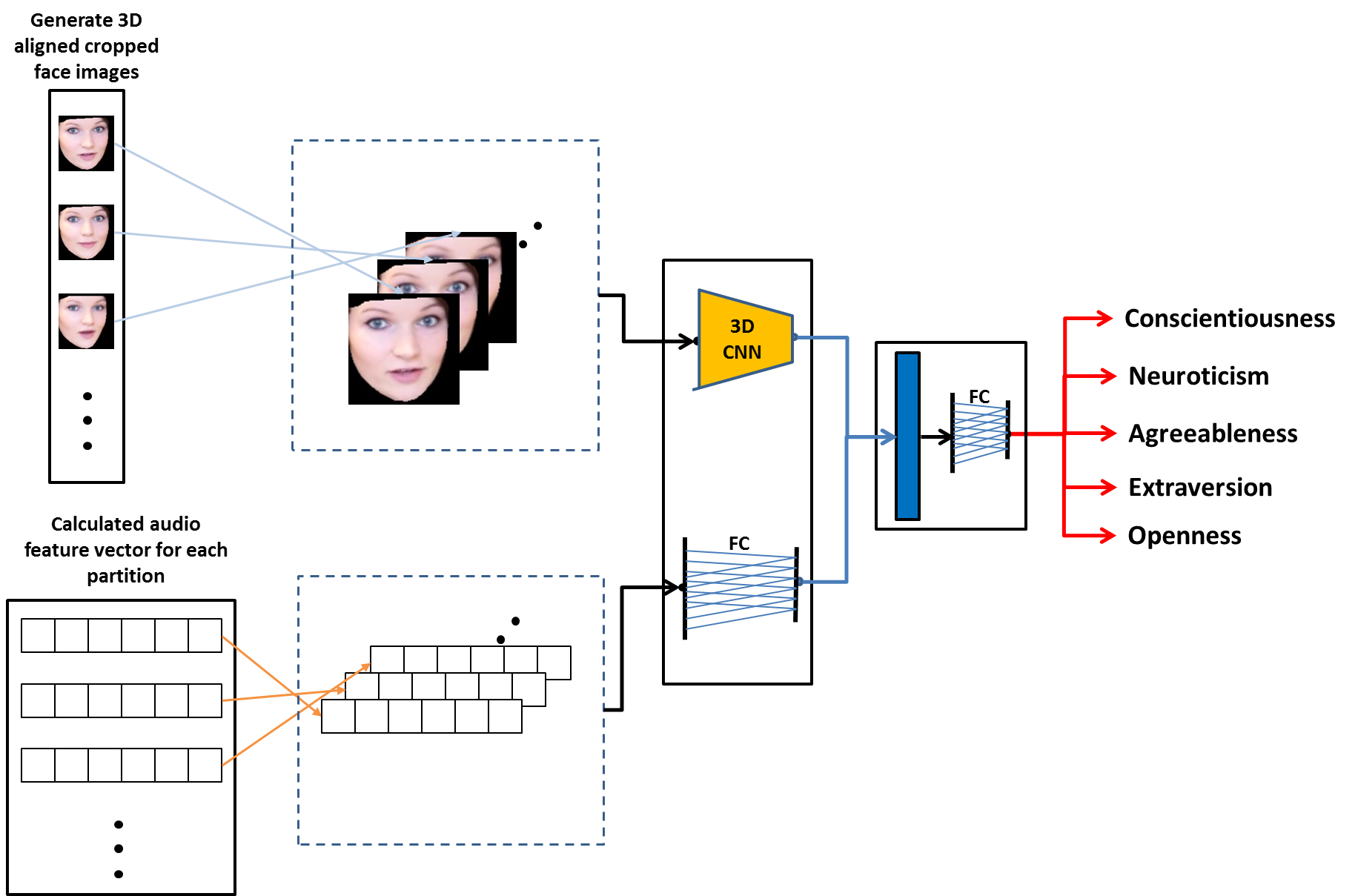}
\caption{Pipeline of 3D-Convolution model\label{convpipe}}
\end{figure}

Our first model is inspired from the work of Ji et. al\cite{ji3dconv}. The architecture is shown in figure \ref{multimodalconv} and the pipeline is demonstrated in figure \ref{convpipe}. The visual data processing branch learns the change in facial expressions from face aligned images using 3D convolution. At first, the 6 face aligned temporally ordered images of size $3\times 112\times 112$ are passed through a 3D convolution layer, followed by a ReLU and a 3D max-pooling layer. The 3D convolution as well as max-pooling are done in a volume comprised of X, Y and t dimensions. The resulting feature maps are in-turn passed through a second set of similar layers of 3D convolution, ReLU and 3D max-pooling but with different kernel sizes (refer to figure \ref{multimodalconv} for details about parameters). This is followed by another layer of 3D convolution, which result in a single feature map of size $1\times21\times21$ which is flattened to a 441 dimensional feature vector. Simultaneously, the audio-data processing branch gets a $6 \times 68$ dimensional feature vector which is reduced to a 100 dimensional vector using a fully connected layer. The feature vectors from audio and visual branches are concatenated and yields a 541 (100 from audio + 441 from visual data) dimensional feature vector, which is then input to a fully connected (FC) layer of 200 nodes and a ReLU layer, followed by another FC layer of 5 nodes which has the activation function as sigmoid. These 5 nodes represent the predicted values of the Big-Five Personality traits.

\subsubsection{LSTM based model:} 
\begin{figure}
\centering
\includegraphics[height=0.3\textheight]{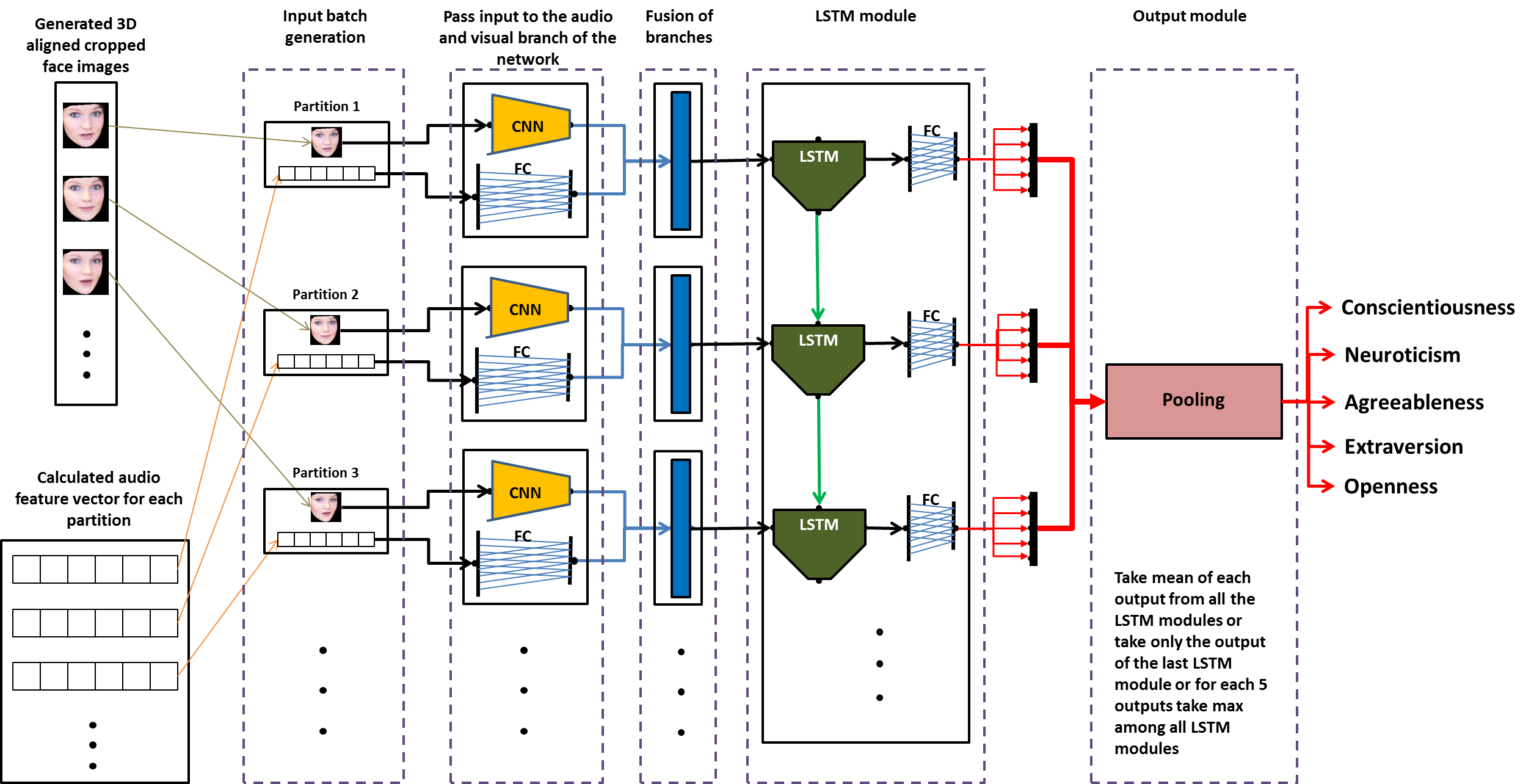}
\caption{Pipeline of LSTM model\label{lstmpipe}}
\end{figure}

We designed our second model to learn the task based on temporal relationship within the input. The architecture and pipeline of the model are shown in figure \ref{multimodallstm} and figure \ref{lstmpipe} respectively. We propose LSTM units to capture the temporal patterns of the input data to predict the personality traits. Each aligned face image is passed through a series of spatial convolution, ReLU and spatial max-pooling layers of varying kernel sizes (refer to figure \ref{multimodallstm} for details about parameters). The generated feature maps are flattened to get 1024 dimensional feature vector and it is connected to a fully connected layer of 128 nodes. Simultaneously, the audio-data (6 feature vectors of 68 dimension) is passed through a 32-node fully connected layer and reduced to 32-dimension. After these steps, the output feature vectors from audio and visual data processing branches are concatenated to yield 6 feature vectors of 160 dimension (32 dim of audio + 128 dim of visual data for each 6 partition) which are still maintained in temporal order. The extracted temporally ordered 6 feature vectors are then passed through an LSTM with output dimension of 128. The LSTM takes $6\times160$ dimensional input and outputs a sequence of 6 $128$-dimensional feature vector. The LSTM generates output for each time step and then, each output is passed through 5 dimensional fully-connected layer with sigmoid activation function. Thus, we get 6 outputs of predicted 5 personality traits. For each personality trait, we average the predicted value, output by all 6 LSTM output units. Thus we get a single prediction value for each of the Big Five personality traits.

\section{Stochastic Training and Testing}
\label{sec:stochastic_training}
According to Psychology research \cite{willis2006first}, it is observed that first impressions of unfamiliar faces can be formed even with exposure times as small as 100-ms.  Their results suggest that predictions made with a 100-ms exposure correlated highly with judgments made in the absence of time constraints, suggesting that small exposure times were sufficient for participants to form an impression. On similar lines, we also hypothesize that deep models can learn effective representations for recognizing first impressions from a few randomly selected frames.

\subsection{Stochastic Training}
Training of the two proposed models is carried out using Stochastic Gradient Optimization (SGD) method. The parameters used for SGD are: learning rate = 0.05, weight decay = $5\times e^{-4}$, momentum = 0.9, batch size = 128, learning rate decay $ = 1\times e^{-4}$.

As mentioned earlier (figure \ref{preprocess}), each raw video file is split into non-overlapping 6 partitions and the audio as well as visual features are extracted from each partition individually. We propose to train the models by using a combined feature set such that we take single face aligned image from each partition, as well as the pre-processed audio features from each partition. Particularly, in video data, since we are only using 1 frame from whole partition, there are multiple combinations of frames from each partition possible for training. Consider there are N partitions \& F frames per partition  and we intend to take a single frame from each partition, hence $F^N$ combinations of frames are possible per video. We assume N as 6 and typically, F is in the range of ${\raise.17ex\hbox{$\scriptstyle\sim$}} 75$ (considering 30 fps and each video of 15 seconds). Training the model with $75^6$ combinations of frames is an overkill. Empirically, we found that training only on several hundreds of combinations (typically {\raise.17ex\hbox{$\scriptstyle\sim$}} 500) is enough for the model to generalize for whole dataset. 

Going with the above explanation, the 6 input frames (single frame from each partition) for model training is selected randomly by keeping temporal ordering in mind. At every epoch, the random selection will yield new input combination for each video.
 This stochastic way of training produces new sample at every epoch and ``regularizes'' the learning effectively, thus increasing the generalization of the model.
\subsection{Testing}
Testing the model also faces the same issue of exponential combination of frames per video. Empirically, we choose to use only a random subset (10 combinations) from total possible combinations and use the average of 10 evaluations as the Personality-traits recognition results. The validation and test results suggest that the model and evaluation method performs significantly better than the other submissions and the LSTM model stood at second place in the Final evaluation phase of competition.
\section{Experiments and Results}
\label{sec:experiments_results}
In this section, we first briefly describe about the dataset and
the evaluation protocol from our experiments. Then we provide the implementation details for our method and discuss
the results.
\subsection{Dataset: Apparent Personality Analysis (APA) - First impressions}
In our validation experiment, we use the ChaLearn LAP 2016 APA dataset provided by the challenge organizers\cite{chalearn1stround1stimpressions}. This dataset has 6000 videos for training with ground truth Personality-traits, 2000 videos for validation without ground truth (performance is revealed on submission of predictions) and 2000 videos for test (Ground truth is not available until the competition is finished). Each video is of length 15 seconds and generally has 30 frames/second. The ground truth consists of fractional scores in the range between 0 to 1 for each of Big-Five Personality traits : Extraversion, Agreeableness, Conscientiousness, Neuroticism, Openness.
 
 \begin{figure}
  \centering
  \includegraphics[scale=0.4]{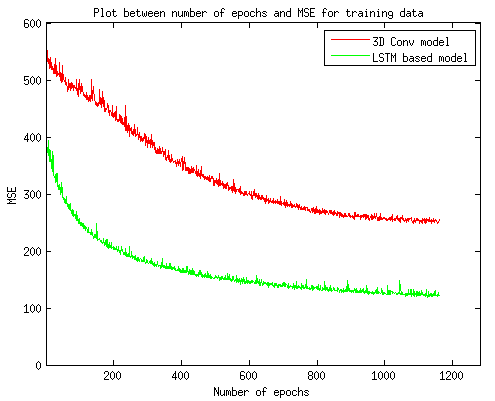}
  \caption{Number of Epochs vs. Mean Squared Error (MSE) for individual models during training phase \label{fig:msecompare}}
  \end{figure}
  
\subsection{Evaluation Protocol}

The evaluation is done in terms of Mean Average Accuracy.\\
The individual personality traits Average Accuracy is calculated as,

\begin{equation}
\text{Average Accuracy}_j = \frac{1}{N} \sum_{i=1}^N(1 - |Target_{ij}-y_{ij}|)
\end{equation}

where j = 1\dots 5, N is the number of total videos, $Target_{ij}$ is the ground truth value for $i^{th}$ video and $j^{th}$ personality-trait, $y_{ij}$ is the predicted value for $i^{th}$ video and $j^{th}$ personality-trait.

The Mean Average Accuracy between the predictions and the ground truth personality-traits values:

\begin{equation}
\text{Mean Average Accuracy} = \frac{1}{m} \sum_{j=1}^m(\text{Average accuracy}_j)
\end{equation}

where m = 5 (the number of Personality Traits).

Note, that the maximum value of the Mean Average Accuracy, as well as Average Accuracy is equal to 1, which represents the best result and the minimum is equal to 0 representing the worst match.

\subsection{Implementation details}
Both of the deep learning models are implemented using Torch\cite{collobert2011torch7} scientific computing framework. The training of 3D convolution based model takes 30 seconds per epoch and LSTM based model takes 3 minutes per epoch on a GeForce GTX Titan Black graphics card. The training of each individual model is done for up-to whole 1 day. We used only the ChaLearn LAP 2016 APA dataset\cite{chalearn1stround1stimpressions} for training. The comparison of mean squared error(MSE) of both models during training is shown in figure \ref{fig:msecompare}. The source code files of both the training and final proposed prediction method are available in github \footnote{refer \href{https://github.com/InnovArul/first-impressions}{https://github.com/InnovArul/first-impressions} for more information} repository.

\subsection{Development phase}
In the development phase of the APA2016 competition\cite{chalearn1stround1stimpressions}, only the training set ground truths were released and the methods were evaluated online by submitting the predictions on the validation videos to a server. The best performance of our models during development phase is shown in Table \ref{validationrankings}. 
\subsection{Test phase}
In the test phase of the APA2016 competition\cite{chalearn1stround1stimpressions}, the testing videos were released. The testing ground truths were kept secret and the teams were invited to submit their results on the testing videos. The organizers announced the final ranking after the test phase. The results are summarized in Table \ref{resultstable}. The proposed LSTM model secured the second place in the leader-board and shown in bold font.

\subsection{Results and Discussion}

The performance of CNN (3D convolution) based model and LSTM model can be seen from learning phase evaluation shown in table \ref{validationrankings}:
\newpage
\begin{longtable}{ l | p{3cm} | l }
	\multicolumn{1}{l}{} &
	\multicolumn{1}{l}{LSTM model} &
	\multicolumn{1}{l}{3D conv. based model} \\ \hline
    Accuracy & \textbf{0.913355} & 0.912473 \\\hline
    Extraversion & 0.914548 & 0.915650 \\\hline
    Agreeableness & 0.915749 & 0.916123	 \\\hline
    Conscientiousness & 0.913594 & 0.908370 \\\hline
    Neuroticism & 0.909814 & 0.909931 \\\hline
    Openness & 0.913069 & 0.912292 \\\hline
\caption{Evaluation during learning phase on ChaLearn LAP 2016  APA : First Impressions challenge \label{validationrankings}}
\end{longtable}


The test phase leader-board standings is shown in the table \ref{resultstable}.

\begin{longtable}{ p{2cm} | p{5cm} | p{2cm} }
	\multicolumn{1}{l}{\textbf{Rank}} &
	\multicolumn{1}{l}{\textbf{Team}} &
	\multicolumn{1}{l}{\textbf{Accuracy}}\\\hline
 1 & NJU-LAMDA & 0.912968 \\\hline
\textbf{2} & \textbf{evolgen (*LSTM model)} & \textbf{0.912063} \\\hline
3 & DCC & 0.910933  \\\hline
4 & ucas & 0.909824 \\ \hline
5 & BU-NKU & 0.909387 \\ \hline
6 & pandora & 0.906275 \\ \hline
7 & Pilab & 0.893602 \\ \hline
8 & Kaizoku & 0.882571 \\ \hline
\caption{Leaderboard of Test-phase on ChaLearn LAP 2016  APA : First Impressions challenge. our entry is with \textbf{bold} \label{resultstable}}
\end{longtable}%

As we noticed from the table \ref{validationrankings}, during learning phase, LSTM based model performs superior to 3D convolution based model. It maybe due to the fact that, LSTM is able to learn better temporal relationships than 3D convolution based approach. Also, the audio-features were not used to define temporal relationship in 3D convolution based model (only 3D face aligned images are used), but LSTM model used both audio and visual features to learn the temporal correspondences, which could have made it perform better. Because of these reasons, we chose LSTM model to be used for test phase: Our method secured second place in ChaLearn LAP 2016: APA challenge\cite{chalearn1stround1stimpressions}.

\section{Conclusions and Future Works}%
\label{sec:conclusions}
In this work, we proposed two deep neural network based models that use audio and visual features for the task of First Impressions Recognition.  These networks mine the temporal patterns that exist in a sequence of frames.  It was also shown that such sequences can be small and selected in a stochastic manner respecting the temporal order.  The proposed methods have been shown to yield excellent performance on the ChaLearn LAP APA2016 Challenge\cite{chalearn1stround1stimpressions}.  As deep neural networks are known for their representation and feature extracting ability, they can be used to learn the optimal representations without having to pre-process the data. Appearance and Pose features can also be explored to see if they improve the performance given by the proposed audio and visual features.

\bibliographystyle{splncs}
\bibliography{0003}

\begin{thebibliography}{10}

\bibitem{wikifirstimpress}
Wikipedia.
\newblock \url{https://en.wikipedia.org/wiki/First\_impression\_(psychology)}
  Definition of psychological term First impression.

\bibitem{willis2006first}
Willis, J., Todorov, A.:
\newblock First impressions making up your mind after a 100-ms exposure to a
  face.
\newblock Psychological science \textbf{17}(7) (2006)  592--598

\bibitem{cowie2001emotion}
Cowie, R., Douglas-Cowie, E., Tsapatsoulis, N., Votsis, G., Kollias, S.,
  Fellenz, W., Taylor, J.G.:
\newblock Emotion recognition in human-computer interaction.
\newblock IEEE Signal processing magazine \textbf{18}(1) (2001)  32--80

\bibitem{cohen2000emotion}
Cohen, I., Garg, A., Huang, T.S.,  et~al.:
\newblock Emotion recognition from facial expressions using multilevel hmm.
\newblock In: Neural information processing systems. Volume~2., Citeseer (2000)

\bibitem{cohen2003facial}
Cohen, I., Sebe, N., Garg, A., Chen, L.S., Huang, T.S.:
\newblock Facial expression recognition from video sequences: temporal and
  static modeling.
\newblock Computer Vision and image understanding \textbf{91}(1) (2003)
  160--187

\bibitem{kim2013deep}
Kim, Y., Lee, H., Provost, E.M.:
\newblock Deep learning for robust feature generation in audiovisual emotion
  recognition.
\newblock In: 2013 IEEE International Conference on Acoustics, Speech and
  Signal Processing, IEEE (2013)  3687--3691

\bibitem{Kessous}
L.Kessous, G.Castellano, G.:
\newblock Multimodal emotion recognition in speech-based interaction using
  facial expression, body gesture and acoustic analysis.
\newblock Journal on Multimodal User Interfaces \textbf{3}(1) (2010)  33--48

\bibitem{Kim}
Zhao, X., Kim, T.K., Luo, W.:
\newblock Unified face analysis by iterative multi-output random forests.
\newblock (2014)

\bibitem{David}
Razuri, J.G., Sundgren, D., Rahmani, R., Moran~Cardenas, A.:
\newblock Automatic emotion recognition through facial expression analysis in
  merged images based on an artificial neural network.
\newblock (2013)

\bibitem{ji3dconv}
Ji, S., Xu, W., Yang, M., Yu, K.:
\newblock 3d convolutional neural networks for human action recognition.
\newblock IEEE transactions on pattern analysis and machine intelligence
  \textbf{35}(1) (2013)  221--231

\bibitem{chalearn1stround1stimpressions}
Lopez, V.P., Chen, B., Places, A., Oliu, M., Corneanu, C., Baro, X., Escalante,
  H.J., Guyon, I., Escalera, S.:
\newblock Chalearn lap 2016: First round challenge on first impressions -
  dataset and results. chalearn looking at people workshop on apparent
  personality analysis.
\newblock In: ECCV Workshop proceedings. (2016)

\bibitem{giannakopoulos2015pyaudioanalysis}
Giannakopoulos, T.:
\newblock pyaudioanalysis: An open-source python library for audio signal
  analysis.
\newblock PloS one \textbf{10}(12) (2015)  e0144610

\bibitem{pyaudioanalysis}
Giannakopoulos, T.:
\newblock pyaudioanalysis.
\newblock \url{https://github.com/tyiannak/pyAudioAnalysis} an open Python
  library that provides a wide range of audio-related functionalities.

\bibitem{pyaudiofeatures}
Giannakopoulos, T.:
\newblock pyaudioanalysis.
\newblock
  \url{https://github.com/tyiannak/pyAudioAnalysis/wiki/3.-Feature-Extraction}
  Features extracted using pyAudioAnalysis.

\bibitem{baltru2016openface}
Baltru, T., Robinson, P., Morency, L.P.,  et~al.:
\newblock Openface: an open source facial behavior analysis toolkit.
\newblock In: 2016 IEEE Winter Conference on Applications of Computer Vision
  (WACV), IEEE (2016)  1--10

\bibitem{openface}
Baltru, T., Robinson, P., Morency, L.P.,  et~al.:
\newblock Openface.
\newblock \url{https://github.com/TadasBaltrusaitis/OpenFace} a state-of-the
  art open source tool intended for facial landmark detection, head pose
  estimation, facial action unit recognition, and eye-gaze estimation.

\bibitem{collobert2011torch7}
Collobert, R., Kavukcuoglu, K., Farabet, C.:
\newblock Torch7: A matlab-like environment for machine learning.
\newblock In: BigLearn, NIPS Workshop. Number EPFL-CONF-192376 (2011)

\end{thebibliography}
\end{document}